\documentclass[journal]{IEEEtran}
\usepackage{makecell}
\usepackage[shortlabels]{enumitem}
\usepackage{algorithm}
\usepackage{algorithmic}
\usepackage{amsmath}      
\usepackage{amssymb}      
\usepackage{hyperref}
\usepackage{graphicx}%
\usepackage{multirow}%
\usepackage{amsmath,amssymb,amsfonts}%
\usepackage{amsthm}%
\usepackage{mathrsfs}%
\usepackage{xcolor}%
\usepackage{textcomp}%
\usepackage{manyfoot}%
\usepackage{booktabs}%
\usepackage{listings}%
\usepackage{soul}
\usepackage{marvosym}
\usepackage{url}
\usepackage{float}
\newtheorem{example}{Example}%

\begin{document}
\title{CSAI: Conditional Self-Attention Imputation for Healthcare Time-series}

\author{Linglong Qian,
        Joseph Arul Raj,
        Hugh Logan-Ellis,
        Ao Zhang,
        Yuezhou Zhang,
        Tao Wang,
        Richard JB Dobson,
        Zina Ibrahim

 \thanks{Manuscript received Nov 4, 2024; revised Aug 10, 2025 and Oct 10, 2025; accepted Dec 13. LQ, YZ, TW, RD \& ZI are supported by the NIHR Biomedical Research Centre (BRC) at the Maudsley. ZI and RD are further supported the University College London Hospitals BRC. JAR, HLE \& AZ are supported by the DRIVE-Health EPSRC Centre for Doctoral Training.
}

\thanks{\textbf{Corresponding author:} Zina Ibrahim, zina.ibrahim@kcl.ac.uk. LQ, JAR, HLE, AZ, YZ, TW, RD and ZI are with the Department of Biostatistics and Health Informatics, King's College London, London UK. }

}

 \maketitle

 \begin{abstract}
We introduce the Conditional Self-Attention Imputation (CSAI) model, a novel recurrent neural network architecture designed to address imputation challenges in multivariate time series derived from hospital electronic health records (EHRs). CSAI introduces key novelties specific to EHR data: \textit{a)} attention-based hidden state initialisation to capture both long- and short-range temporal dependencies, \textit{b)} domain-informed temporal decay to mimic clinical recording patterns, and \textit{c)} a non-uniform masking strategy that models non-random missingness. Comprehensive evaluation across four EHR benchmark datasets demonstrates CSAI's effectiveness compared to state-of-the-art architectures in data restoration and downstream tasks. CSAI is integrated into PyPOTS, an open-source Python toolbox for partially observed time series. This work significantly advances the state of neural network imputation applied to EHRs by more closely aligning algorithmic imputation with clinical realities.
\end{abstract}


\section{Introduction}\label{sec1}

Multivariate time series from electronic health records (EHR) are essential for deriving patient-specific insights \cite{healthcare}, yet their complexity poses significant challenges. EHR data suffer from extensive missingness due to clinical and administrative workflows that create irregular recording patterns \cite{mitra2023learning} — e.g., heart rate is monitored routinely, whilst white blood cell counts are ordered only in specific scenarios, such as suspected infection. Moreover, EHR structure generates strong correlations between feature values and their missingness patterns over time, e.g. hypertension's link to kidney disease leads to high correlations between blood pressure measurements and creatinine levels, both in terms of values and recording patterns \cite{correlation}. These irregularities result in over 50\% of the EHR data missing not at random \cite{wells2013strategies}, with substantial variation in missingness patterns across tasks and datasets \cite{wells2013strategies}, creating challenges for imputation algorithms \cite{oursurvey,jensen2012mining,survey}.

Recurrent neural networks (RNNs) have shown promise in handling EHR missingness by efficiently modeling sequential dependencies~\cite{grud,cao2018brits,yoon2017multi}. Among these, BRITS (Bidirectional Recurrent Imputation for Time Series)\cite{cao2018brits} stands out in performance \cite{oursurvey}. However, general imputation architectures, including BRITS, overlook the unique nature of medical data collection. As a result, they struggle to capture long-range EHR correlations\cite{jensen2012mining} (e.g. long-term HbA1c levels in diabetic patients) and are not focused on capturing non-random EHR misingness~\cite{schafer2002missing}, significantly oversimplifying the temporal and cross-sectional dependencies of real EHR time series.

To address these issues, we developed CSAI, a bi-directional RNN using BRITS as a backbone, extending it to better suit EHR data characteristics through the following contributions: 

\begin{enumerate}[label=\arabic*., nosep, leftmargin=*, align=left]


\item Improved hidden state initialisation using a transformer-based conditional self-attention to capture long-term dynamics, complementing the RNN's short-term dynamics.


\item A domain-informed \emph{temporal decay} function reflecting clinical recording patterns, where each feature's decay factor adjusts its associated attention mechanism for more precise, feature-specific temporal representation.

\item CSAI is integrated with a non-uniform masking strategy to selectively reflect the naturally structured patterns of interdependencies within the dataset across time and features. 
\end{enumerate}

EHR missingness patterns convey clinically relevant information. By aligning with those, CSAI improves both imputation quality and downstream predictive model performance.

\section{Related Work}\label{sec2}

EHR Time-series imputation progressed substantially through deep learning~\cite{wang2024deep}. GRU-D~\cite{grud} pioneered RNN imputation by estimating missing values via decaying memory of past observations, assuming that older values are less relevant. M-RNN~\cite{yoon2017multi} and BRITS~\cite{cao2018brits} extended GRU-D by capturing bidirectional dynamics and cross-feature correlations. BRITS consistently demonstrates strong performance across datasets, whilst M-RNN does not generalise beyond specific domains.

Beyond RNNs, transformer-based time-series imputers employ self-attention to capture global contextual relationships in temporal missingness patterns~\cite{transformerstimeseriessurvey}.
With medical time-series, however, transformers imposes high computational demands and require adjustments to preserve sequential integrity \cite{du2023saits}.

Other architectures include convolutional neural networks (CNNs)~\cite{kravchik2018detecting,timesnet}, which capture local or spatial patterns but often struggle to maintain temporal consistency. Graph neural networks (GNNs)~\cite{tsignn} model inter-variable dependencies, yet constructing effective graph structures from irregular time series remains challenging. Generative frameworks, including VAEs~\cite{hivae}, GANs~\cite{luo2019e2gan}, and diffusion-based models~\cite{tashiro2021csdi}, learn complex data distributions but face training instability, scalability and leakage issues. While theoretically appealing, these methods lack tailored mechanisms for clinical data~\cite{oursurvey}.

Regardless of the approach, existing models predominantly rely on random masking during training \cite{oursurvey}, which creates oversimplified scenarios that do not reflect missingness patterns observed in actual EHRs. Motivated by these gaps, we propose CSAI to retain the robust sequential learning capabilities of RNNs, while addressing key EHR-specific challenges.

\section{Terminology and Background}\label{sec3}

\subsection{Incomplete Multivariate Time-series Representation} \label{data definition}
\begin{figure}[ht]
\centering
\includegraphics[width=1
\linewidth]{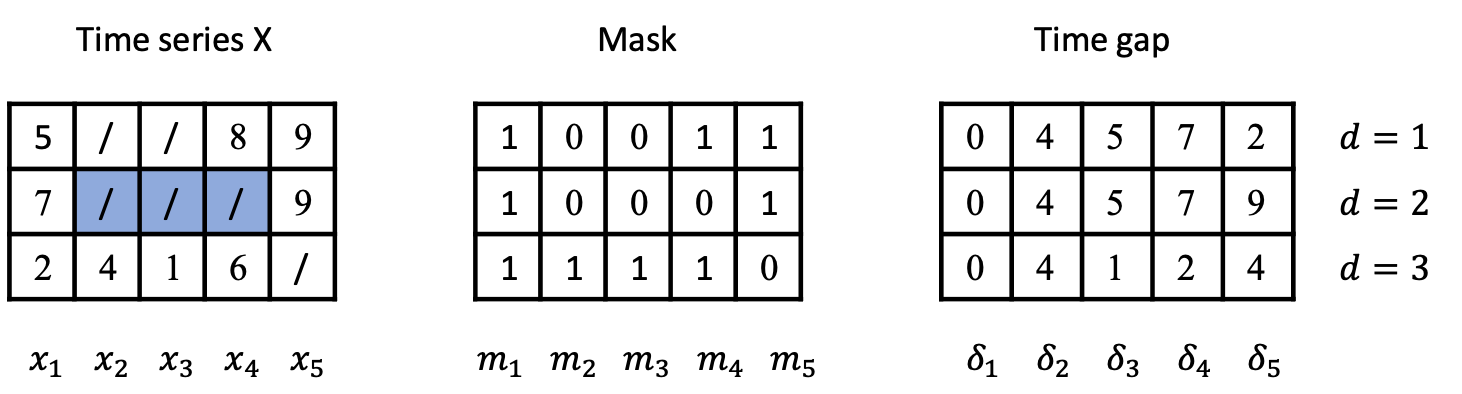}
\caption{An example of multivariate time-series. Observations $\boldsymbol{x_{1-5}}$ in time-stamps $\boldsymbol{s_{1-5}}=0,\,4,\,5,\,7,\,9$. Feature $d_2$ was missing during $\boldsymbol{s_{2-4}}$, the last observation took place at $\boldsymbol{s_1}$. Hence, $\boldsymbol{\delta_5^2}=\boldsymbol{t_5-t_1} =9-0=9$.}
\label{fig:1}
\end{figure}

We represent a multivariate time series as a matrix  $\boldsymbol{X} \in \mathbb{R}^{T\times D}$. $\boldsymbol{X}=\{{\boldsymbol{x_{1}},..., \boldsymbol{x_{T}}}\}$ comprises $T$ observation vectors, $\boldsymbol{x_{t}} \in \mathbb{R}^{1 \times D}$ of $D$ features observed at timestamp $s_t$. Two derived matrices describe missingness (Fig.\ref{fig:1}). The mask matrix $\boldsymbol{M} \in \mathbb{R}^{T\times D}$ indicates whether each element of $\boldsymbol{X}$ is observed:
\begin{equation} 
{m_{t}^d} = \begin{cases} 0, &\text{if} \ x_{t}^d \ \text{is missing}\\ 1,&\text{otherwise} \end{cases} 
\end{equation}

Furthermore, because the time between consecutive observations may vary, we denote the time gaps at each time step by an additional component $\boldsymbol{\delta^d_t} \in \mathbb{R}^{T \times D}$, encoding the gap between two successive observed values for a feature $d$, providing an additional indicator of temporal context.  
\begin{equation} 
{\delta_{t}^d} = \begin{cases} {s_t - s_{t-1} + \delta_{t-1}^d} & \text{if} \ t > 1,\ m_{t}^d = 0\\
{s_t - s_{t-1}} & \text{if} \ t > 1, \ m_{t}^d =1 \\ 0 &\text{if} \ t = 1 
\end{cases} 
\end{equation}

\subsection{Overview of the BRITS Backbone}\label{brits}
BRITS exploits cross-sectional and temporal correlations in multivariate time series $\boldsymbol{X}$ through two dedicated components, a fully connected regression module and a recurrent module. Missing values within observation $\boldsymbol{x_t}$ are managed via corresponding masking and time-gap vectors $\boldsymbol{m_t}$ and $\boldsymbol{\delta_t}$. Because $\boldsymbol{x_t}$ may contain missing values, the BRITS components cannot directly ingest it. Instead, BRITS uses the notion of \emph{decay} to derive estimates for missing cells. Temporal decay dictates that the strengths of the correlations are inversely related to the time-gap $\boldsymbol{\delta_t}$ and is formalised by the decay factor $\gamma_{t} \in (0,1]$, Eq.\eqref{eq:decay}. $\gamma_{t}$ is used to transform $\boldsymbol{h}_{t-1}$ to a \emph{decayed} hidden state, $\boldsymbol{\hat{h}}_{t-1} $, Eq.\eqref{eq:decayedstate}, which is used to find the historical estimation $\boldsymbol{\hat{x}_t} $, of $\boldsymbol{x_t}$, Eq.\eqref{eq:regression}. Using masking vector $\boldsymbol{m_t} \in \boldsymbol{M}$, $\boldsymbol{\hat{x}_t}$ is then used to replace the missing values of $\boldsymbol{x_t}$, yielding the \emph{complement vector} $\boldsymbol{x_t^{h}} $, which embeds temporal missingness patterns to be fed into subsequent BRITS components, Eq.\eqref{eq:complement}.
\begin{align}
\label{eq:h0} \boldsymbol{h_0} &= 0 \\
\label{eq:decay} \gamma_{th} &= \exp\left(-\max(0,\boldsymbol{W_{\gamma}} \boldsymbol{\delta_t} + \boldsymbol{b_{\gamma}})\right)\\ 
\label{eq:decayedstate} \boldsymbol{\hat{h}_{t-1}} &= \boldsymbol{h_{t-1}} \odot \gamma_{t} \\
\label{eq:regression} \boldsymbol{\hat{x}_t} &= \boldsymbol{W_x}\boldsymbol{\hat{h}_{t-1}} + \boldsymbol{b_x}\\
\label{eq:complement} \boldsymbol{x_t^{h}} &= \boldsymbol{m_t }\odot \boldsymbol{x_t} + (1-\boldsymbol{m_t}) \odot \boldsymbol{\hat{x}_t} 
\end{align}

BRITS explores cross-sectional correlations within an observation through a fully-connected layer, generating $\boldsymbol{x_t^{f}}$, a feature-based approximation of missing values, Eq.\eqref{eq:full}. The decay concept extends to the feature space, resulting in a learnable factor $\hat{\beta}{_t}$, Eq.\eqref{eq:decayspatial}-\eqref{eq:beta}. This integration produces the imputed matrix $\boldsymbol{x_t^c}$, effectively combining observed and imputed data (Eq.\eqref{eq:compbeta}-\eqref{eq:imputed}). The final step updates the hidden state via its RNN component, leveraging indicators to learn functions of past observations, Eq.\eqref{eq:rnn}. Bidirectional dynamics address slow convergence using backwards information. 

\begin{align}
\label{eq:full} \boldsymbol{x_t^{f}} &= \boldsymbol{W_z x_t^{h}}+\boldsymbol{b_z} \\
\label{eq:decayspatial} \gamma_{tf} &= \exp\left(-\max(0,\boldsymbol{W_{\gamma f}} \boldsymbol{\delta_t} + \boldsymbol{b_{\gamma f}})\right)\\ 
\label{eq:beta} \hat{\beta}_{t} &= \sigma(\boldsymbol{W_\beta} [\gamma_{tf} \circ \boldsymbol{m_t}] +\boldsymbol{b_\beta}) \\
\label{eq:compbeta} \boldsymbol{\hat{x}_t^c} &= \beta_t \odot \boldsymbol{x_t^{f}} + (1-\beta_t) \odot \boldsymbol{x_t^{h}} \\ 
\label{eq:imputed} \boldsymbol{x_t^c} &= \boldsymbol{m_t }\odot \boldsymbol{x_t} + (1-\boldsymbol{m_t}) \odot \boldsymbol{\hat{x}_t^c} \\
\label{eq:rnn} \displaystyle \boldsymbol{h_t} &= \sigma(\boldsymbol{W_t \hat{h}_{t-1}} + \boldsymbol{U_h}[\boldsymbol{x_t^c}\circ \boldsymbol{m_t}]+ \boldsymbol{b_h})
\end{align}

\section{Methodology}\label{sec4}

We now describe the modifications of the BRITS architecture to incorporate \textbf{ a)} a domain-informed temporal decay functionality, \textbf{b)} a transformer-based hidden state initialisation capturing long-range correlations, and \textbf{c) }a novel non-uniform masking strategy to explicitly model non-random EHR missingness. The section concludes CSAI's learning framework.

\subsection{EHR-Tailored BRITS Adaptations}
\noindent\textbf{Domain-informed Temporal Decay:}
The BRITS decay function Eq.\eqref{eq:decay} is strictly dependent on temporal proximity, dynamically adjusting the contribution of a past observation to a missing value based on the length of the time gap between the two. Although this mechanism captures the intuition that more recent observations carry greater diagnostic value, it overlooks domain-specific discrepancies, where different features follow distinct recording frequencies due to clinical practices. 

\begin{example} 
Let feature  \(f_1\): heart rate (HR) \& \(f_2\): systolic blood pressure (SBP). HR is typically monitored more frequently than SBP. An observation $\boldsymbol{x_t}$ has both HR and SBP values missing. The time gap vector $\boldsymbol{\delta_t}$ shows $\boldsymbol{\delta^{f_1}_t} = 2$ (i.e, the last HR recording occurred 2 time units ago) and $\boldsymbol{\delta^{f_2}_t} = 7$ (i.e., the previous SBP recording occurred 7 time units ago). Based on temporal decay, BRITS would incorrectly assign greater weight to the last HR observation, overlooking domain-specific recording patterns where SBP retains significant diagnostic importance  critical for imputation despite the longer time gap.

\end{example}

Our proposed decay mechanism prioritises recent observations while accounting for the natural variability in healthcare data collection. In addition to using $\boldsymbol{\delta^d_t}$, we modify the decay function to incorporate \emph{the expected time gap} $\boldsymbol{\tau}$ between two recordings. \(\tau_d\) is the median of the time intervals between successive recordings of a feature \(d\) in the entire dataset. This adjustment allows the decay function to adapt to recording patterns, ensuring that features like SBP, which are recorded less frequently, still carry appropriate weight during imputation. The new decay factor \(\gamma_t^d\) \(d\) at time \(t\) is computed as:

\begin{equation} \label{attention_weight} 
    \gamma_t^d = \exp(-\max(0, W_\gamma (\delta_t^d - \tau_d) + b_\gamma))
\end{equation}

\(\delta_t^d\) is the time gap since the last observation of feature \(d\),
\(\tau_d\) is \(d\)'s median time gap, and
 \(W_\gamma\) and \(b_\gamma\) are learnable parameters. This formulation ensures that the decay factor peaks when the time gap \(\tau_t^d\) closely matches the expected gap \(\tau_d\), and declines as the difference between \(\delta_t^d\) and \(\tau_d\) increases, ensuring that observations within their expected time gap contribute more strongly to the imputation process.

\begin{example} 
Continuing from the previous scenario, examining the dataset reveals median time gaps \(\boldsymbol{\tau_1} = 2\) for $f_1$ (HR) and \(\boldsymbol{\tau_2} = 10\) for $f_2$ (SBP), reflecting that HR is routinely monitored more frequently than SBP in clinical practice. The model can account for the different recording frequencies by leveraging these median time gaps. Since the last observed values for both features fall within their respective median time gaps, the model assigns comparable importance to both past recordings when imputing missing values, preserving the clinical relevance of the less frequently measured SBP. 
\end{example}

\vspace{.2cm}
\noindent\textbf{Attention-based Hidden State Initialisation: }The hidden states of BRITS' recurrent component are not generated through the raw input. Instead, they receive incomplete data with missingness indicators for imputation, i.e. $\boldsymbol{\hat{h}}$ replaces $\boldsymbol{h}$ (Eq.\eqref{eq:decayedstate}). However, BRITS' initial hidden states are initialised to zero (Eq. \ref{eq:h0}), causing the model to rely solely on internal parameters to estimate initial missing values (Eq. \ref{eq:regression}), ignoring prior observations. This is problematic where early data points are crucial to understanding patient trends, and failure to incorporate them can lead to inaccurate imputations.

\begin{example} For a given patient, HR has been steadily increasing, indicating deterioration, but a monitoring gap causes missing initial values. With zero hidden-state initialisation, BRITS fails to capture the upward trend by disregarding information from later measurements, continuously stacking errors, and potentially misrepresenting the patient's condition as stable when urgent intervention is required.
\end{example}

To overcome this, we use the last observed data point and a decay attention mechanism to generate an initial hidden state conditional distribution $\boldsymbol{q}(\boldsymbol{h}_{\text{init}}|\boldsymbol{x}_{\text{last\_obs}}, \gamma_t^d ))$ within the model distribution $\boldsymbol{p_\theta}(\boldsymbol{x_t})$, providing a richer starting point. Instead of applying the decay factor directly to the previous hidden state as in BRITS (Eqs. \eqref{eq:decayedstate} and \eqref{eq:beta}), we use the decay factor to modulate an attention mechanism, capturing better long-range and feature-specific dynamics. First, at each time step \(s_t\), the last observation \(\boldsymbol{x_{\text{last\_obs}}} \in \mathbb{R}^{T \times D_{\text{feature}}}\) and decay factor \(\gamma_t^d\) are projected and encoded to capture their temporal position:

\begin{align}
\boldsymbol{x^{'}_{\text{last\_obs}}} &= \text{PosEncoder}(\text{InputProj}(\boldsymbol{x}_{\text{last\_obs}})) \\
\gamma^{'}_t &= \text{PosEncoder}(\text{InputProj}(\gamma_t^d))
\end{align}

Then the transformed input representations are concatenated and fed into a Transformer encoder, which captures both long-range dependencies and feature-specific interactions:

\begin{equation}
\boldsymbol{C}_{\text{in}} = \text{Concat}(\boldsymbol{x^{'}_{\text{last\_obs}}}, \gamma^{'}_t)
\end{equation}
\begin{equation}
\boldsymbol{C}_{\text{out}} = \text{LN}(\text{FFN}(\text{LN}(\text{MSA}(\boldsymbol{C}_{\text{in}}))))
\end{equation}

The transformer output is then passed through 1D convolutions to adjust dimensions and initialise the hidden state:

\begin{figure*}[ht]
\centering
\includegraphics[width=0.8\linewidth]{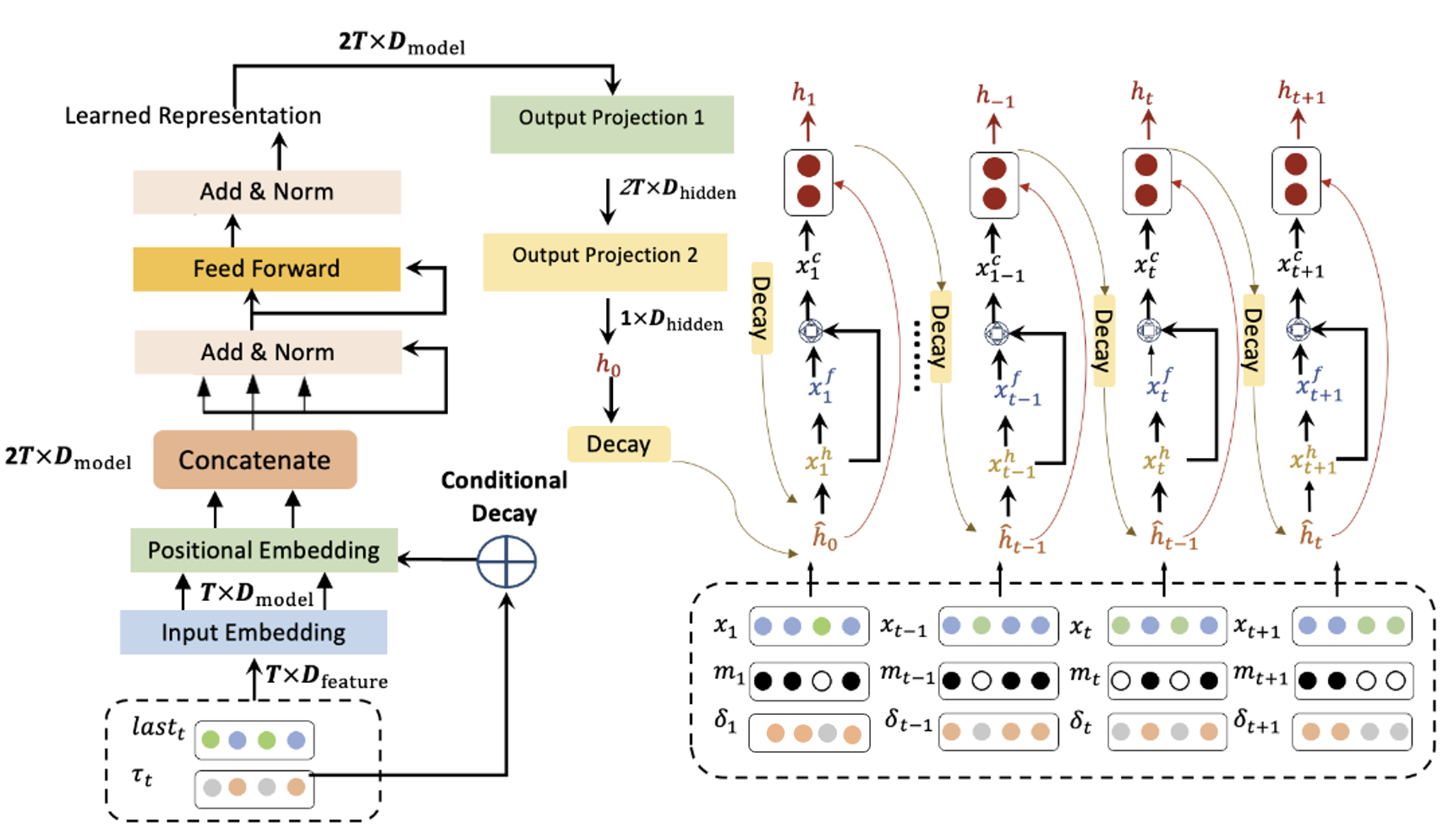}
\caption{The CSAI architecture, beginning with an input embedding layer, followed by a positional embedding to capture time dependencies. Embeddings are processed through multi-head attention, normalisation, and feed-forward layers. The output initialises hidden states for subsequent recurrent layers.}
\label{fig:CSAI}
\end{figure*}

\begin{align}
\boldsymbol{H_1} &= \text{Conv1D}_1(\boldsymbol{C}_{\text{out}} \boldsymbol{W_1} + \boldsymbol{b_1}) \label{Conv1}\\
\boldsymbol{h}_{\text{init}} &= \text{Conv1D}_2(\boldsymbol{H_1} \boldsymbol{W_2} + \boldsymbol{b_2}) \label{Conv2}
\end{align}

Eq.\eqref{Conv1} transforms $\boldsymbol{C}_{\text{out}}$ from $\mathbb{R}^{2L \times d_{\text{model}}}$ to $\mathbb{R}^{2L \times d_{\text{hidden}}}$ and produces $\boldsymbol{H_1}$, Eq.\eqref{Conv2} further scales $\boldsymbol{H_1}$ to generate the initialised hidden state $\boldsymbol{h}_{\text{init}}$, allowing the model to capture variations in feature-recording, providing a robust foundation for subsequent steps of the CSAI architecture (Fig. \ref{fig:CSAI}).

\subsection{Non-Uniform Masking Strategy}\label{subsec2}

Our masking algorithm diverges from traditional approaches by leveraging the dataset's missingness distribution to generate masking probabilities, incorporating clinical recording patterns into our masking process to create a more realistic missingness model. Our algorithm relies on two factors:


\begin{enumerate}[wide, labelindent=0pt]

\item \noindent \textbf{Missingness Distribution} $\boldsymbol{P_{\text{dist}}}(d)$: This reflects the likelihood of masking a feature based on its missingness patterns across similar or neighbouring observations.

\item \noindent \textbf{Adjustment Factor} $R_{\text{factor}}(d)$: dynamically adjusts a feature $d$'s masking probability based on observation frequency to avoid overfitting while maximising the utility of limited data.

\end{enumerate}

 For a given feature $d$, the non-uniform masking probability $\boldsymbol{P_{\text{nu}}}(d)$ is determined as follows:
\begin{align}
\boldsymbol{R_{\text{factor}}}(d|U,I) &= \boldsymbol{F}(d, U, I) \label{eq:mask}\\
\boldsymbol{P_{\text{nu}}}(d) &= \boldsymbol{R_{\text{factor}}}(d|U,I) \times \boldsymbol{P_{\text{dist}}}(d) \label{eq:maskingprobability}
\end{align}
 $U$ is a predetermined masking rate: the percentage of ground truths masked during training. $I$ is a weighting parameter. 
   $\boldsymbol{R_{\text{factor}}}(d|U,I)$ is the adjustment factor for feature $d$, conditioned by $U$ and $I$.
    $\boldsymbol{P_{\text{dist}}}(d)$ is $d$'s missingness distribution.

\begin{algorithm}
\begin{algorithmic}
\STATE \textbf{Input:} $\boldsymbol{X}$ with $D$ features, $U$, $I$
\STATE \textbf{Output:} Masked Dataset $\boldsymbol{X_M}$

\FOR{each feature $d$ in $D$}
    \STATE $\boldsymbol{P_{\text{dist}}}(d) \gets \text{compute}(\boldsymbol{x^d})$
    \STATE $\boldsymbol{R_{\text{factor}}}(d) \gets f(U,I)$
    \STATE $\boldsymbol{P_{\text{nu}}}(d) \gets \boldsymbol{R_{\text{factor}}}(d) \times \boldsymbol{P_{\text{dist}}}(d)$
\ENDFOR
\STATE $U \gets f(\boldsymbol{P_{\text{nu}}}, \boldsymbol{X})$; $\boldsymbol{X_M} \gets \boldsymbol{P_{\text{nu}}} \times \boldsymbol{X}\;$
\end{algorithmic}
\caption{\emph{\textbf{non-uniform-mask}}}\label{algo2}
\end{algorithm}

\begin{figure}[ht]
\centering
\includegraphics[width=
\linewidth]{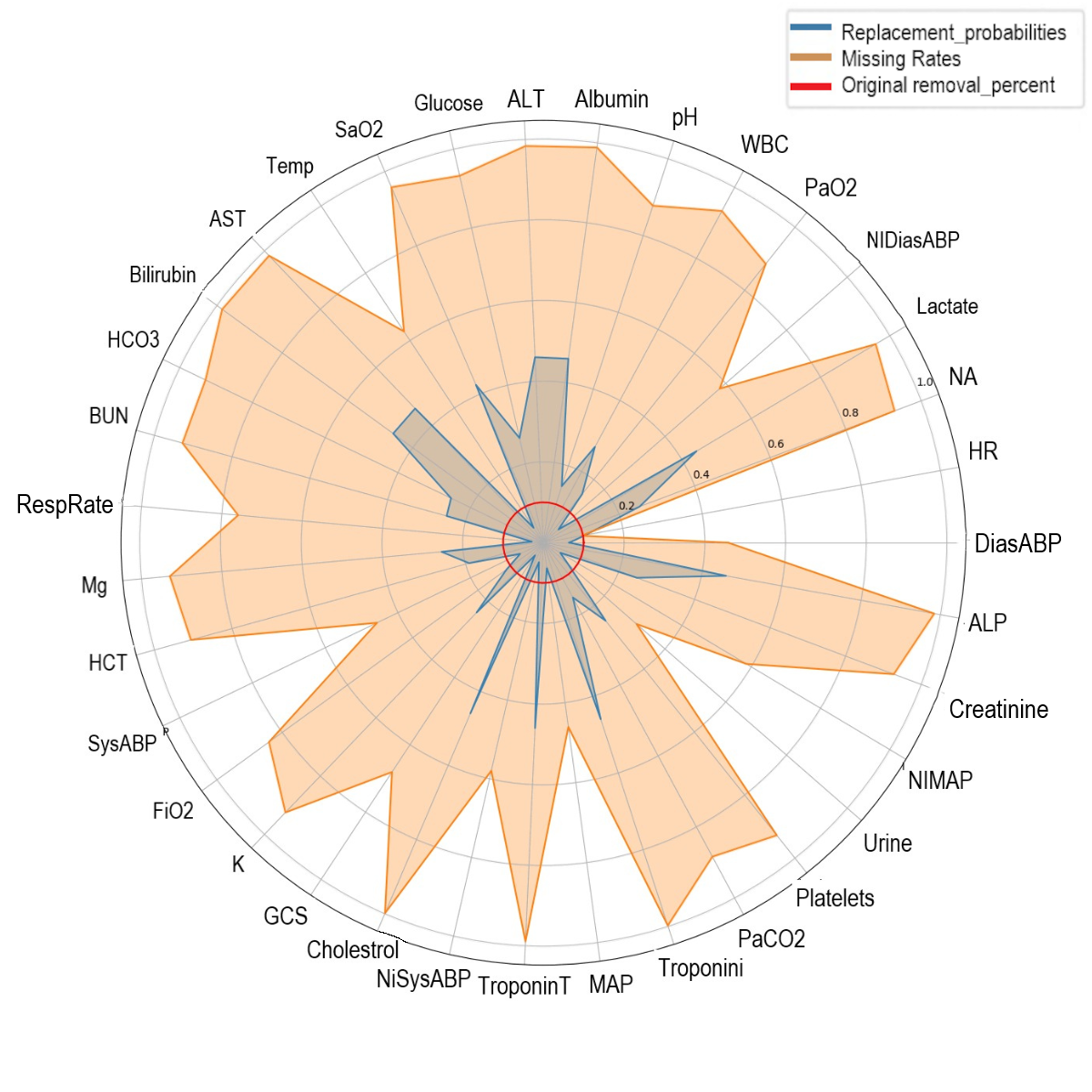}
\caption{Non-Uniform EHR masking patterns across 40 features: \textbf{Red Circle:} original uniform masking percentage; \textbf{Blue Area}: non-uniform masking probabilities; \textbf{Orange Area: }feature missingness rates, highlighting the alignment between actual missingness and resulting non-uniform masking probabilities.}
\label{fig:nonuniformmasking}
\end{figure}

The overall masking proportions are adjusted to ensure consistency with the masking rate $U$, while retaining each feature's non-uniform characteristics. The adjustment factor (Eq. \ref{eq:mask}) scales down the masking probability for features with naturally high missingness (Eq. \ref{eq:maskingprobability}) as illustrated in Fig.\ref{fig:nonuniformmasking} to prevent overfitting to sparse observations by masking sparse features less often, promoting learning from limited data. Modulation strength is controlled by the weighting parameter $I$, tuned to achieve optimal trade-off between imputation accuracy and generalisation (see ablation study, Section \ref{sec:experimental_analysis}). Our algorithm produces the masked matrix $\boldsymbol{X_M} $.

\subsection{Learning}

CSAI is trained in an unsupervised manner by non-uniformly masking non-missing values and learning to impute them. CSAI's training iterates over mini-batches of $T$ time steps of the input data. Our imputation loss $\mathcal{L}_{imp}$ minimises the reconstruction error $\ell_{\text{obs}}$ between the imputed and ground truth vectors $\boldsymbol{x}$ and $\boldsymbol{\hat{x}}$, while maintaining consistency loss $\ell_{\text{con}}$ between the forward and backward imputed estimates $\overrightarrow{\hat{\boldsymbol{x}}}$ and $\overleftarrow{\hat{\boldsymbol{x}}}$ of our bi-directional RNN as follows:


\begin{align}
\mathcal{L}_{imp} = \frac{1}{|\mathcal{B}|} \sum_{i \in \mathcal{B}} \sum_{t \in T_i} \lambda  \Big[ & \boldsymbol{M}_{it} \odot \ell_{\text{obs}}(\boldsymbol{x}_{it}, \hat{\boldsymbol{x}}_{it}) \nonumber \\
& + (1 - \boldsymbol{M}_{it}) \odot \ell_{\text{con}}(\overrightarrow{\hat{\boldsymbol{x}}_{it}}, \overleftarrow{\hat{\boldsymbol{x}}_{it}}) \Big]
\end{align}

Where $\mathcal{B}$ is the mini-batch size, $T$ is the set of time steps over which the loss function is applied, $\boldsymbol{M}$ is the masking matrix, and $\lambda$ is a hyperparameter that balances the two loss terms. Both $\ell_{obs}$ and $\ell_{cons}$ use MAE. 
 Since CSAI can be used as an end-to-end pipeline to perform imputation and prediction, we define prediction loss using binary cross-entropy, Eq. \ref{eq:predictionloss}, and the combined loss $\mathcal{L}_{\text{combined}}$, Eq. \ref{eq:combinedloss}.
\begin{eqnarray}\mathcal{L}_{\text{pred}} = - \sum_{n=1}^{N} y^{(n)} \log(\hat{y}^{(n)}) \label{eq:predictionloss} \\
\mathcal{L}_{\text{combined}} = \alpha \mathcal{L}_{\text{imp}} + \beta \mathcal{L}_{\text{pred}} \label{eq:combinedloss}
\end{eqnarray}
Where $\alpha$ and $\beta$ are preset parameters representing the weight of the respective loss component on $\mathcal{L}_{\text{combined}} $. Using CSAI solely as an imputer is done by setting $\beta$ to zero. While BRITS uses an LSTM for its recurrent component, we use a Gated Recurrent Unit (GRU) in CSAI due to their computational efficiency (see our ablation study, Section \ref{sec:experimental_analysis}).

\section{Experimental Evaluation}\label{sec6}

\subsection{Datasets}
We used three widely-used healthcare benchmarks with varying characteristics and missingness distributions (Table \ref{tab:datasets}). We reproduced the available benchmarks, skipping \textbf{NAN}-removal steps to retain the data's original missingness:

\begin{table}[t]
\centering
\caption{Dataset Characteristics: \textbf{Size}:samples $\times$ features. \textbf{Average baseline missingness} across features. \textbf{Feature correlation}: average feature-wise Pearson correlation coefficient. \textbf{(static; categorical)} number of static and categorical features.}
\small
\begin{tabular}{l@{\hskip 4pt}c@{\hskip 4pt}c@{\hskip 4pt}c@{\hskip 4pt}c}
\toprule
\textbf{Dataset} & \textbf{Size} & \textbf{Missing} & \textbf{Corr.} & \textbf{(Static; Cat.)} \\
\midrule
eICU         & 30,680×20   & 40.53\% & 0.14 & (6; 3) \\
MIMIC\_59    & 21,128×59   & 61\%    & 0.17 & (0; 0) \\
Physionet    & 3,997×35    & 51\%    & 0.12 & (0; 0) \\
\bottomrule
\end{tabular}
\label{tab:datasets}
\end{table}

\begin{enumerate}[wide, labelindent=0pt]
\item \textbf{eICU}\cite{pollard2018eicu}: a publicly-available database of anonymised intensive care unit (ICU) records from over 200 hospitals. We followed the only benchmark extract available for eICU \cite{eicubenchmark}.

\item \textbf{MIMIC-III} \cite{johnson2016mimic}: a public database of more than 40,000 patients. We followed a well-cited benchmark \cite{harutyunyan2019multitask}. 

\item \textbf{PhysioNet Challenge 2012 dataset} \cite{physionetchallenge}: a public medical benchmark of 4000 48-hour hospital stays.

\end{enumerate}

\subsection{Experimental Design} 

Our experiments \textbf{benchmark} CSAI's performance against the state of the art and evaluate CSAI's individual components: 


\noindent\textbf{Experiment I: } compares CSAI with the four best-performing RNN models using the three benchmarks and 5\%, 10\% and 20\% masking ratios. Here, we compare CSAI with BRITS \cite{cao2018brits}, GRU-D \cite{grud}, V-RIN \cite{mulyadi2021uncertainty}, and M-RNN \cite{yoon2017multi}. Here, we use the original source code obtained from each model's respective \texttt{GitHub} repository. To ensure fairness, we employed our non-uniform masking as a pre-processing step for all models. 

\noindent\textbf{Experiment II:} is a large-scale comparison with different neural imputation architectures using the Physionet dataset and 10\% masking. This experiment was conducted using PyPOTS\footnote{https://pypots.com/about/}\cite{pypots}, an open-source Python toolkit providing standardised access to 29 imputation algorithms, including CSAI and benchmark datasets, including Physionet. We compare CSAI with Transformer, CNN, GNN, RNN and diffusion-based imputers.
We note that although a comparison with generative models through $E^2$GAN \cite{luo2019e2gan} would have provided valuable insight, particularly given its lack of quantitative comparison with BRITS, $E^2$GAN's available implementation is incompatible with our setup and is not part of PyPOTS. 

\noindent\textbf{Experiment III:} is an \textbf{ablation study} carried out on Physionet. Here, we: a) incrementally evaluate the contribution of CSAI's components to impuation performance, b) evaluate non-uniform masking across training, validation, and test partitions to examine its effect on handling missing data, and c) examine the weighting parameter, which controls feature emphasis in the non-uniform masking strategy and allows feature representation fine-tuning during training, assessing its impact on CSAI's imputation and classification performance.

\subsection{Experimental Setup} 

\noindent{\textbf{Environment:}} We used an HPC node equipped with NVIDIA A100 40GB, running Ubuntu 20.04.6 LTS (Focal Fossa), using Python 3.8.16 . To ensure reproducibility, the full package details, frozen Anaconda environment, pre-processing scripts, model implementation and hyperparameter search configurations are available on the project's \texttt{GitHub} repository.

\noindent{\textbf{Dataset Pre-processing and Missingness Simulation:}} For all datasets, we masked 5\%, 10\% and 20\% cells in addition to the missingness already present in the datasets. These masked cells have known ground truths and will form the basis for performance evaluation. To more effectively simulate EHR scenarios, our first set of experiments uses our non-uniform masking as a pre-processing step to capture common EHR missingness patterns in the masked cells. In our PyPOTs experiments, we used the package's PyGrinder toolbox\footnote{https://pypots.com/ecosystem/\#PyGrinder}, which implements structured missingness patterns \cite{mitra2023learning}.

\noindent{\textbf{Base Architecture \& Hyper-parameter Optimisation:}} 
To ensure fairness and distinguish CSAI's architectural contributions from the choice of its recurrent cell, we evaluated BRITS with both its original LSTM implementation (BRITS\_LSTM) and a GRU variant we implemented, mirroring CSAI's base (BRITS\_GRU). To eliminate bias from manual tuning, all hyperparameter optimisations were conducted systematically through the standardised PyPOTS interface.

\noindent{\textbf{Training:}} Except for the PhysioNet dataset, which already contains separated time series samples, all other datasets are split into training, validation, and test sets. Randomly, we selected 10\% of each dataset for validation and another 10\% for testing, training the models on the remaining data.  We used the Adam optimiser and set the number of RNN hidden units to 108 for all models. The batch size is 64 for PhysioNet and 128 for the other datasets. A 5-fold cross-validation method was implemented to evaluate the models. 

\noindent{\textbf{Downstream Task Design:}} We further perform a classification task to predict the in-hospital mortality outcome provided in the benchmarks. To demonstrate the flexibility of the model, we varied the methodologies when implementing the classifiers in our experiments. In \textbf{Experiment I}, classification was performed end-to-end by adding a classification layer to each architecture, utilising the hidden states from the imputation network to feed into the classification layer. In \textbf{Experiment II}, the imputation models were used to generate complete datasets, which were subsequently fed into two different classifiers for comparison: an XGBoost and an RNN classifier. 

\vspace{.2cm}

\subsection{Experimental Results} \label{sec:experimental_analysis}

\noindent{\textbf{Experiment I: Comparison with RNN Models:}} \label{sec:performance}
Our evaluation is shown in Tables \ref{tab:task_I} and \ref{tab:task_C}. Table \ref{tab:task_I} shows the imputation performance, where CSAI consistently outperforms other models in all data sets and masking ratios (5\%, 10\%, and 20\%). The best performance for all models is observed in eICU and MIMIC\_59, which, despite high missingness rates, offer a large number of training samples. Physionet, while having the lowest baseline missingness rate, provides significantly fewer samples, limiting model performance across the board. The table also shows that the performance gap between CSAI and other models widens as the masking ratio increases, leading to conditions of high data loss. This is particularly pronounced in the highly-dimensional MIMIC\_59 at the 20\% missingness ratio. For V-RIN, BRITS, and BRITS\_GRU, the MAE is relatively stable across masking ratios but remains consistently higher than CSAI. GRUD and MRNN show notably higher MAE values, especially at higher masking ratios.

\begin{table}[htbp]
\centering
\caption{Imputation performance. Bold Highlight: lowest MAE.}
\begin{tabular}{r|ccc}
\toprule

 & \textbf{5\% masking} & \textbf{10\% masking} & \textbf{20\% masking} \\
\midrule
\multicolumn{4}{c}{\textbf{eICU (MAE)}} \\
\midrule
\textbf{V-RIN} & 0.2416 ± 0.015 & 0.2425 ± 0.013 & 0.2521 ± 0.019 \\
\textbf{BRITS\_LSTM} & 0.1669 ± 0.014 & 0.1705 ± 0.020 & 0.1768 ± 0.009 \\
 \textbf{BRITS\_GRU} & 0.1723 ± 0.010 & 0.1712 ± 0.019 & 0.1769 ± 0.013 \\
\textbf{GRUD} & 0.2227 ± 0.018 & 0.2256 ± 0.010 & 0.2309 ± 0.020 \\
\textbf{MRNN} & 0.4704 ± 0.015 & 0.4799 ± 0.017 & 0.5007 ± 0.020 \\
\textbf{CSAI} & \textbf{0.1597 ± 0.017} & \textbf{0.1615 ± 0.011} & \textbf{0.1664 ± 0.015} \\
\midrule
\multicolumn{4}{c}{\textbf{MIMIC\_59 (MAE)}} \\
\midrule

\textbf{V-RIN} & 0.1546 ± 0.007 & 0.1382 ± 0.017 & 0.3369 ± 0.010 \\
\textbf{BRITS\_LSTM} & 0.1519 ± 0.018 & 0.1402 ± 0.009 & 0.3404 ± 0.019 \\
 \textbf{BRITS\_GRU} & 0.1479 ± 0.016 & 0.1419 ± 0.015 & 0.3417 ± 0.017 \\
\textbf{GRUD} & 0.3045 ± 0.012 & 0.2870 ± 0.014 & 0.4867 ± 0.017 \\
\textbf{MRNN} & 0.3057 ± 0.013 & 0.2834 ± 0.012 & 0.4719 ± 0.015 \\
\textbf{CSAI} & \textbf{0.1312 ± 0.009} & \textbf{0.1129 ± 0.008} & \textbf{0.3098 ± 0.014} \\
\midrule
\multicolumn{4}{c}{\textbf{PhysioNet (MAE)}} \\
\midrule
\textbf{V-RIN} & 0.2616 ± 0.015 & 0.2737 ± 0.010 & 0.2999 ± 0.018 \\
\textbf{BRITS\_LSTM} & 0.2563 ± 0.013 & 0.2676 ± 0.017 & 0.2872 ± 0.014 \\
 \textbf{BRITS\_GRU} & 0.2513 ± 0.012 & 0.2622 ± 0.011 & 0.2829 ± 0.018 \\
\textbf{GRUD} & 0.4941 ± 0.015 & 0.4978 ± 0.020 & 0.5095 ± 0.018 \\
\textbf{MRNN} & 0.5467 ± 0.013 & 0.5565 ± 0.014 & 0.5723 ± 0.017 \\
\textbf{CSAI} & \textbf{0.2460 ± 0.014} & \textbf{0.2575 ± 0.017} & \textbf{0.2748 ± 0.019} \\
\bottomrule
\end{tabular}
\label{tab:task_I}
\end{table}

Table \ref{tab:task_C} follows similar patterns, demonstrating CSAI's strong classification performance. CSAI achieves the highest AUC scores across various masking ratios and a slight degradation in performance as masking ratio increases to 20\%. Although a similar stability is observed in BRITS and V-RIN, the AUCs are consistently lower than CSAI's. The difference between CSAI's AUCs and those of other models is also highest in Physionet, where training data is limited. 


\begin{table}[htb]
\centering
\caption{Classification performance. Bold Highlight: Highest AUC.}
\begin{tabular}{r|ccc}
\toprule
 & \textbf{5\% masking} & \textbf{10\% masking} & \textbf{20\% masking} \\
\midrule
\multicolumn{4}{c}{\textbf{eICU (AUC)}} \\
\midrule

\textbf{V-RIN} & 0.8877 ± 0.012 & 0.8842 ± 0.013 & 0.8846 ± 0.015 \\
\textbf{BRITS\_LSTM} & 0.8867 ± 0.011 & 0.8852 ± 0.013 & 0.8857 ± 0.012 \\
\textbf{BRITS\_GRU} & 0.8894 ± 0.012 & 0.8886 ± 0.015 & 0.8861 ± 0.011 \\
\textbf{GRUD} & 0.8649 ± 0.013 & 0.8646 ± 0.014 & 0.8580 ± 0.011 \\
\textbf{MRNN} & 0.8779 ± 0.011 & 0.8763 ± 0.014 & 0.8734 ± 0.015 \\
\textbf{CSAI} & \textbf{0.8895 ± 0.012} & \textbf{0.8898 ± 0.011} & \textbf{0.8879 ± 0.015} \\
\midrule
\multicolumn{4}{c}{\textbf{MIMIC\_59 (AUC)}} \\
\midrule

\textbf{V-RIN} & 0.8328 ± 0.010 & 0.8331 ± 0.012 & 0.8273 ± 0.014 \\
\textbf{BRITS\_LSTM} & 0.8282 ± 0.010 & 0.8278 ± 0.012 & 0.8241 ± 0.013 \\
\textbf{BRITS\_GRU} & 0.8319 ± 0.010 & 0.8307 ± 0.012 & 0.8269 ± 0.013 \\
\textbf{GRUD} & 0.8277 ± 0.012 & 0.8264 ± 0.014 & 0.8230 ± 0.012 \\
\textbf{MRNN} & 0.8211 ± 0.012 & 0.8171 ± 0.011 & 0.8128 ± 0.013 \\
\textbf{CSAI} & \textbf{0.8352 ± 0.014} & \textbf{0.8337 ± 0.013} & \textbf{0.8311 ± 0.012} \\
\midrule
\multicolumn{4}{c}{\textbf{PhysioNet (AUC)}} \\
\midrule

\textbf{V-RIN} & 0.8343 ± 0.011 & 0.8292 ± 0.010 & 0.8255 ± 0.015 \\
\textbf{BRITS\_LSTM} & 0.8221 ± 0.014 & 0.8118 ± 0.012 & 0.8218 ± 0.015 \\
\textbf{BRITS\_GRU} & 0.8068 ± 0.014 & 0.8193 ± 0.013 & 0.8065 ± 0.015 \\
\textbf{GRUD} & 0.7899 ± 0.011 & 0.7765 ± 0.013 & 0.7699 ± 0.015 \\
\textbf{MRNN} & 0.8012 ± 0.014 & 0.7995 ± 0.013 & 0.7940 ± 0.015 \\
\textbf{CSAI} & \textbf{0.8647 ± 0.014} & \textbf{0.8592 ± 0.013} & \textbf{0.8372 ± 0.015} \\
\bottomrule
\end{tabular}
\label{tab:task_C}
\end{table}

\begin{table*}[ht]
\caption{Physionet Performance Metrics (Imputation, Computational, and Downstream Classification) using PyPOTS and 10\% masking.}
\centering
\fontsize{8.5}{10.5}\selectfont
\setlength{\tabcolsep}{3pt}
\begin{tabular}{c|c|c|ccccc|cc}
\toprule
\multicolumn{2}{c|}{\centering\textbf{}} & \centering\textbf{Imputation} & \multicolumn{5}{c|}{\centering\textbf{Computational Metrics (Imputation)}} & \multicolumn{2}{c}{\centering\textbf{Classification (ROC-AUC)}} \\
\cmidrule(lr){3-3} \cmidrule(lr){4-8} \cmidrule(lr){9-10}
\centering\textbf{Model Type} & \centering\textbf{Model Name} & \centering\textbf{MAE} & \textbf{Avg Best} & \textbf{No.} & \textbf{Peak GPU} & \textbf{Training } & \textbf{Inference} & \textbf{XGB } & \textbf{RNN } \\
\textbf{} & \textbf{} & \textbf{} & \textbf{Epoch} & \textbf{Params.} & \textbf{Memory (MiB)} & \textbf{Time (s)} & \textbf{Time (s)} & \textbf{ Classifier} & \textbf{ Classifier} \\
\midrule
\midrule
\multirow{9}{*}{\rotatebox{90}{\textbf{Transformers}}} 
& \textbf{iTransformer} & $0.3698 \pm 0.01$ & $130.4$ & $6.85M$ & $470.4$ & $241.1 \pm 64.8$ & $0.11 \pm 0.02$ & $0.835 \pm 0.00$ & $0.697 \pm 0.11$ \\
& \textbf{SAITS} & $0.2653 \pm 0.02$ & $41.4$ & $44.30M$ & $1233.2$ & $173.0 \pm 65.4$ & $0.35 \pm 0.00$ & $0.841 \pm 0.00$ & $0.699 \pm 0.03$ \\
& \textbf{ETSformer} & $0.3735 \pm 0.01$ & $92.2$ & $9.79M$ & $768$ & $217.7 \pm 59.3$ & $0.23 \pm 0.00$ & $0.818 \pm 0.00$ & $0.723 \pm 0.06$ \\
& \textbf{PatchTST} & $0.2965 \pm 0.01$ & $60.2$ & $644K$ & $2254$ & $276.5 \pm 78.3$ & $0.32 \pm 0.01$ & $0.848 \pm 0.00$ & $\mathbf{0.749 \pm 0.02}$ \\
& \textbf{Crossformer} & $0.3686 \pm 0.16$ & $44$ & $1.22M$ & $976$ & $96.6 \pm 42.8$ & $0.17 \pm 0.00$ & $0.824 \pm 0.00$ & $0.733 \pm 0.06$ \\
& \textbf{Informer} & $0.2961 \pm 0.00$ & $70.4$ & $4.26M$ & $348$ & $97.8 \pm 16.1$ & $0.09 \pm 0.00$ & $0.848 \pm 0.00$ & $0.704 \pm 0.06$ \\
& \textbf{Autoformer} & $0.4193 \pm 0.01$ & $33$ & $14.27M$ & $1015.2$ & $88.0 \pm 22.6$ & $0.26 \pm 0.05$ & $0.777 \pm 0.00$ & $0.583 \pm 0.02$ \\
& \textbf{Pyraformer} & $0.2965 \pm 0.01$ & $37.4$ & $3.74M$ & $274.4$ & $46.1 \pm 15.6$ & $0.11 \pm 0.00$ & $0.836 \pm 0.00$ & $0.716 \pm 0.07$ \\
& \textbf{Transformer} & $0.2709 \pm 0.01$ & $23.8$ & $13.70M$ & $496$ & $40.0 \pm 10.2$ & $0.09 \pm 0.00$ & $0.842 \pm 0.00$ & $0.732 \pm 0.04$ \\
\midrule
\multirow{5}{*}{\rotatebox{90}{\textbf{RNN}}} 
& \textbf{BRITS\_LSTM} & $0.2917 \pm 0.00$ & $99$ & $181K$ & $450.4$ & $569.5 \pm 26.9$ & $2.40 \pm 0.00$ & $0.834 \pm 0.00$ & $0.682 \pm 0.04$ \\
& \textbf{BRITS\_GRU} & $0.2914 \pm 0.00$ & $78.8$ & $142K$ & $450.4$ & $530.7 \pm 38.6$ & $2.39 \pm 0.01$ & $0.827 \pm 0.00$ & $0.689 \pm 0.07$ \\
& \textbf{MRNN} & $0.6767 \pm 0.00$ & $3.8$ & $1.59M$ & $1980$ & $168.7 \pm 8.9$ & $2.17 \pm 0.01$ & $0.776 \pm 0.00$ & $0.620 \pm 0.01$ \\
& \textbf{GRUD} & $0.4147 \pm 0.00$ & $14$ & $136K$ & $40$ & $117.8 \pm 3.5$ & $0.65 \pm 0.01$ & $0.855 \pm 0.00$ & $0.691 \pm 0.09$ \\
& \textbf{CSAI} & $\mathbf{0.2401 \pm 0.00}$ & $39.4$ & $4.77M$ & $577.2$ & $277.4 \pm 91.8$ & $1.26 \pm 0.01$ & $\mathbf{0.860 \pm 0.00}$ & $0.702 \pm 0.04$ \\
\midrule
\multirow{3}{*}{\rotatebox{90}{\textbf{CNN}}} 
& \textbf{TimesNet} & $0.4250 \pm 0.01$ & $32$ & $64.91M$ & $2882$ & $246.7 \pm 53.8$ & $0.40 \pm 0.01$ & $0.809 \pm 0.00$ & $0.743 \pm 0.02$ \\
& \textbf{MICN} & $0.3738 \pm 0.03$ & $90.2$ & $226.66M$ & $5551.6$ & $372.2 \pm 205.6$ & $0.22 \pm 0.00$ & $0.810 \pm 0.00$ & $0.710 \pm 0.04$ \\
& \textbf{SCINet} & $0.3449 \pm 0.00$ & $59.6$ & $5.68M$ & $310.8$ & $81.7 \pm 12.2$ & $0.09 \pm 0.00$ & $0.823 \pm 0.00$ & $0.686 \pm 0.06$ \\
\midrule
\multirow{5}{*}{\rotatebox{90}{\textbf{GNNs}}} 
& \textbf{StemGNN} & $0.3167 \pm 0.01$ & $151.8$ & $1.80M$ & $593.2$ & $341.0 \pm 62.4$ & $0.17 \pm 0.00$ & $0.842 \pm 0.00$ & $0.563 \pm 0.12$ \\
& \textbf{FreTS} & $0.3172 \pm 0.01$ & $80.6$ & $1.87M$ & $3714$ & $218.9 \pm 29.5$ & $0.30 \pm 0.00$ & $0.846 \pm 0.00$ & $0.693 \pm 0.05$ \\
& \textbf{Koopa} & $0.4170 \pm 0.01$ & $16.8$ & $168K$ & $44$ & $106.5 \pm 47.1$ & $0.15 \pm 0.00$ & $0.846 \pm 0.00$ & $0.550 \pm 0.08$ \\
& \textbf{DLinear} & $0.3714 \pm 0.00$ & $16$ & $222K$ & $90$ & $9.0 \pm 0.3$ & $0.04 \pm 0.00$ & $0.834 \pm 0.00$ & $0.697 \pm 0.04$ \\
& \textbf{FiLM} & $0.4559 \pm 0.00$ & $20.4$ & $7.1K$ & $34$ & $37.7 \pm 4.9$ & $0.12 \pm 0.00$ & $0.828 \pm 0.00$ & $0.598 \pm 0.09$ \\
\midrule
\multirow{3}{*}{\rotatebox{90}{\textbf{Diffus.}}} 
& \textbf{CSDI} & $0.2516 \pm 0.00$ & $113.8$ & $1.53M$ & $4352$ & $4810.9 \pm 815.4$ & $387.29 \pm 0.20$ & $0.853 \pm 0.00$ & $0.553 \pm 0.11$ \\
& \textbf{US-GAN} & $0.3114 \pm 0.00$ & $13.8$ & $3.71M$ & $562$ & $440.1 \pm 67.6$ & $2.37 \pm 0.01$ & $0.849 \pm 0.00$ & $0.747 \pm 0.04$ \\
& \textbf{GP-VAE} & $0.4318 \pm 0.00$ & $97.6$ & $502K$ & $166.4$ & $84.8 \pm 12.5$ & $1.76 \pm 0.05$ & $0.835 \pm 0.00$ & $0.603 \pm 0.10$ \\
\bottomrule
\end{tabular}%
\label{tab:pypots}
\end{table*}

\noindent{\textbf{Experiment II: Large-scale Comparison Using PyPOTs: }}
Table \ref{tab:pypots} shows imputation and classification results and computational performance for 24 neural imputation models. This experiment uses the imputed data generated by each model as input to two classifiers: XGBoost and RNN. Note that non-uniform masking was only available to CSAI as it is part of its PyPOTS implementation, but at the time of the writing, it was not yet integrated PyPOT's callable interface. 

Imputation MAE is lowest for CSAI by a large margin. Overall, the XGBoost classifier achieved the best performance (mean AUC: XGBoost: 0.828, RNN: 0.675), confirming that better imputation enables better classification using simpler models. With XGBoost, using \textbf{CSAI} as imputer produced the best performance. Since XGBoost is a non-temporal model, it has clearly benefited from CSAI's ability to encode sequential dependencies (across features and time) as informative features that complement XGBoost’s strengths in feature-based learning. For the RNN classifier, Transformer, CNN and diffusion imputers produced the highest AUCs, showing that their respective architectural inductive biases complement the RNN classifier's ability to learn sequential dynamics. Despite this, CSAI ranked highly under the RNN classifier: a) it produced competitive AUC performance and achieved the highest performance among RNN-based imputers, and b) the imputation MAEs of Transformer, CNN, and diffusion models were substantially higher than CSAI's, indicating that while architectural synergy enabled strong classification, these imputers performed poorly at accurately filling missing data.

The table also highlights CSAI's computational efficiency. CSAI exhibits a manageable model size (4.77M parameters) and efficient training profile, especially compared to the heavy Transformer models. CSAI's efficiency stems from its faster convergence, enabling earlier stopping during training (CSAI: 39.4 epochs; BRITS: 78.8 epochs) and is primarily attributable to CSAI's architecture: a) richer initialisation reduces the iterations required for the model to stabilise, b) clinically-informed learning of relevant temporal decay dynamics, and c) training on more realistic missingness patterns and mitigating overfitting to sparse observations using non-uniform masking.

\begin{figure}[htbp]
\centering
\includegraphics[width=\linewidth]{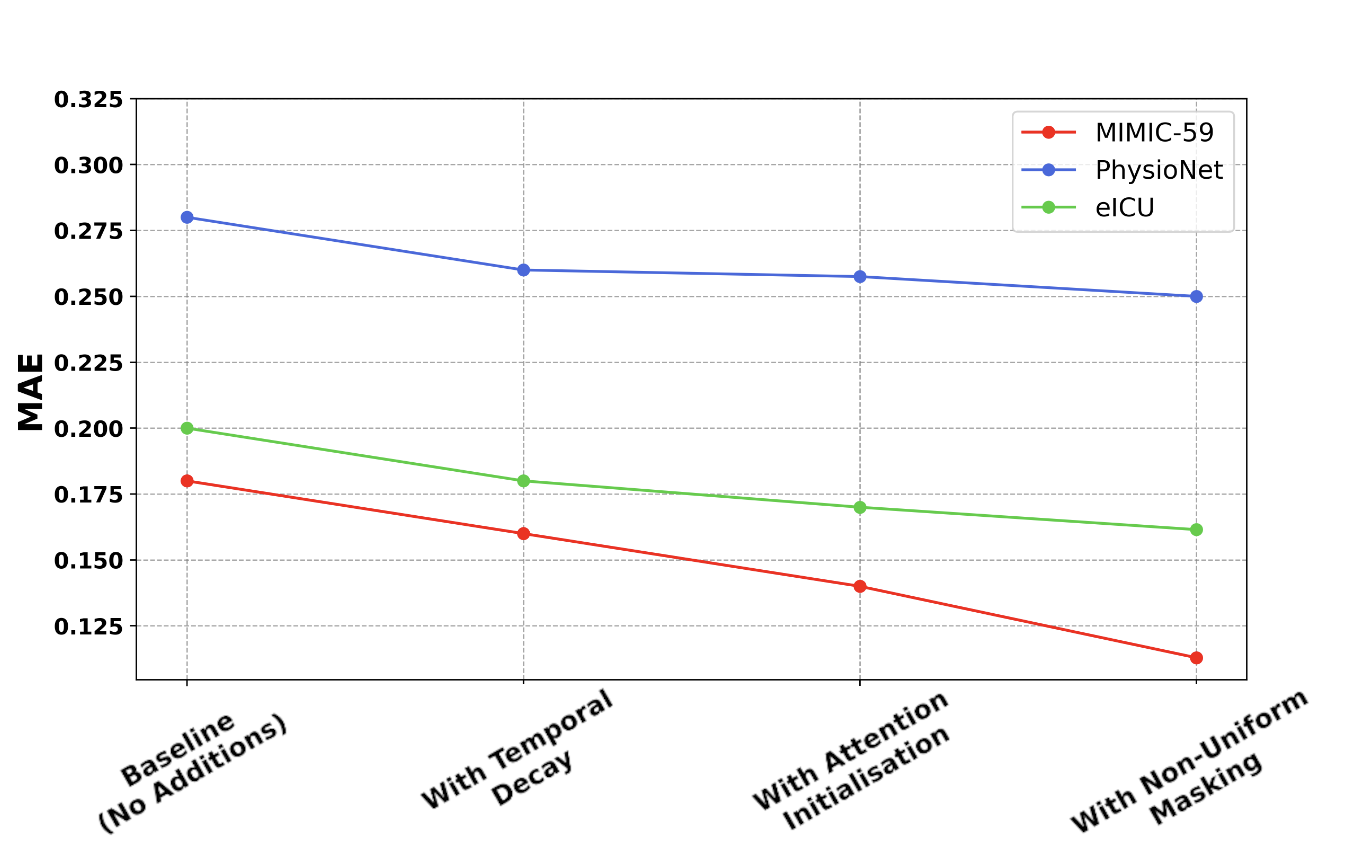}
\caption{MAE descreases as CSAI components are incrementally added.}
\label{fig:ablation1}
\end{figure}

\noindent{\textbf{Ablation Study: }}Figure \ref{fig:ablation1} demonstrates a clear progression as CSAI's baseline MAE improves with every additional component, highlighting the impact of domain-informed temporal decay, attention-based initialisation, and non-uniform masking. The final configuration (full model) achieves the lowest MAE across all data sets, indicating that each component plays a distinct role in enhancing imputation accuracy.

We further evaluate non-uniform masking by using the masking strategy on different training, validation, and test sets combinations and examining the number of training epochs required to achieve the reported performance for each. For objective evaluation, we conducted all experiments using the BRITS baseline model. The results (Table \ref{tab:masking_exps}) demonstrate that consistently applying non-uniform masking across all data partitions (training, validation and testing) yields the best performance, suggesting that the masking strategy effectively adjusts the representation of different features to optimally leverage the data distribution, enhancing the model's ability to handle the inherent heterogeneity of the dataset.

\begin{table}[htbp]
\centering
\caption{Effect of non-uniform masking on performance. 'All' configuration: non-uniform masking is applied to all subsets.}
\resizebox{\linewidth}{!}{
\begin{tabular}{cccc|ccc}
\toprule
\multirow{2}[4]{*}{\textbf{Model}} & \multirow{2}[4]{*}{\textbf{Masking}} & \multicolumn{2}{c|}{\textbf{Imputation}} & \multicolumn{3}{c}{\textbf{Classification}} \\
\cmidrule{3-7}          &       & \textbf{Epoch} & \textbf{MAE} & \textbf{Epoch} & \textbf{MAE} & \textbf{AUC} \\
\midrule
\multirow{6}[2]{*}{\textbf{BRITS}} & All   & 182.2 & \textbf{0.234929} & 28.6  & 0.262997 & \textbf{0.819142*} \\
      & Val\_Test & 187   & 0.235739 & 57    & 0.262268 & 0.816184 \\
      & Test\_only & 218.6 & 0.236001 & 61.6  & 0.265496 & 0.813821 \\
      & Train\_only & 215.8 & 0.266307 & 25.4  & 0.310624 & 0.815686 \\
      & None  & 218.6 & 0.26762 & 61.6  & 0.313257 & 0.81175 \\
      & Val\_only & 187   & 0.268386 & 57    & 0.309332 & 0.818605 \\
\midrule

\bottomrule
\end{tabular}%
}
\label{tab:masking_exps}%
\end{table}%

The non-uniform masking probability used by CSAI for each feature is determined by the parameters $U$ and $I$, in addition to the feature's prior probability as shown in Eq.\eqref{eq:mask}- \eqref{eq:maskingprobability}. We studied the effect of the weighting parameter $I$ on the resulting imputation and classification performance (Table \ref{tab:masking_table}).
An optimal weighting parameter, shown to be around 5, results in the lowest imputation error, suggesting that a balanced representation of features is crucial for accuracy. However, increasing the weighting parameter for classification leads to higher errors and a marginal decrease in the AUC, highlighting that excessive weighting may not uniformly improve performance across different machine learning tasks. These findings reveal the interplay and between feature representation adjustments and task-specific model efficacy, requiring calibration.

\begin{table}[htbp]
\centering
\caption{Impact of weighting parameter $I$ on performance.}

    \begin{tabular}{ccc|ccc}
    \toprule
    \multirow{2}[4]{*}{\textbf{$I$}} & \multicolumn{2}{c|}{\textbf{Imputation}} & \multicolumn{3}{c}{\textbf{Classification}} \\
\cmidrule{2-6}          & \textbf{Epoch} & \textbf{MAE} & \textbf{Epoch} & \textbf{MAE} & \textbf{AUC} \\
    \midrule
    0     & 286   & 0.26215514 & 15.4  & 0.31849296 & 0.8192579 \\
    10    & 294.8 & 0.23159396 & 21.8  & 0.26615049 & 0.8118691 \\
    50    & 293   & 0.24170042 & 16.4  & 0.28670924 & 0.81583396 \\
    100   & 285.8 & 0.26885496 & 14.2  & 0.32437366 & 0.81381879 \\
    150   & 283.8 & 0.29943347 & 16.6  & 0.35346112 & 0.8144707 \\
    200   & 289.2 & 0.33235399 & 15.6  & 0.39360851 & 0.81173828 \\
    \bottomrule
    \end{tabular}%

\label{tab:masking_table}%
\end{table}%

\begin{figure}[htbp]
\centering
\includegraphics[width=\linewidth]{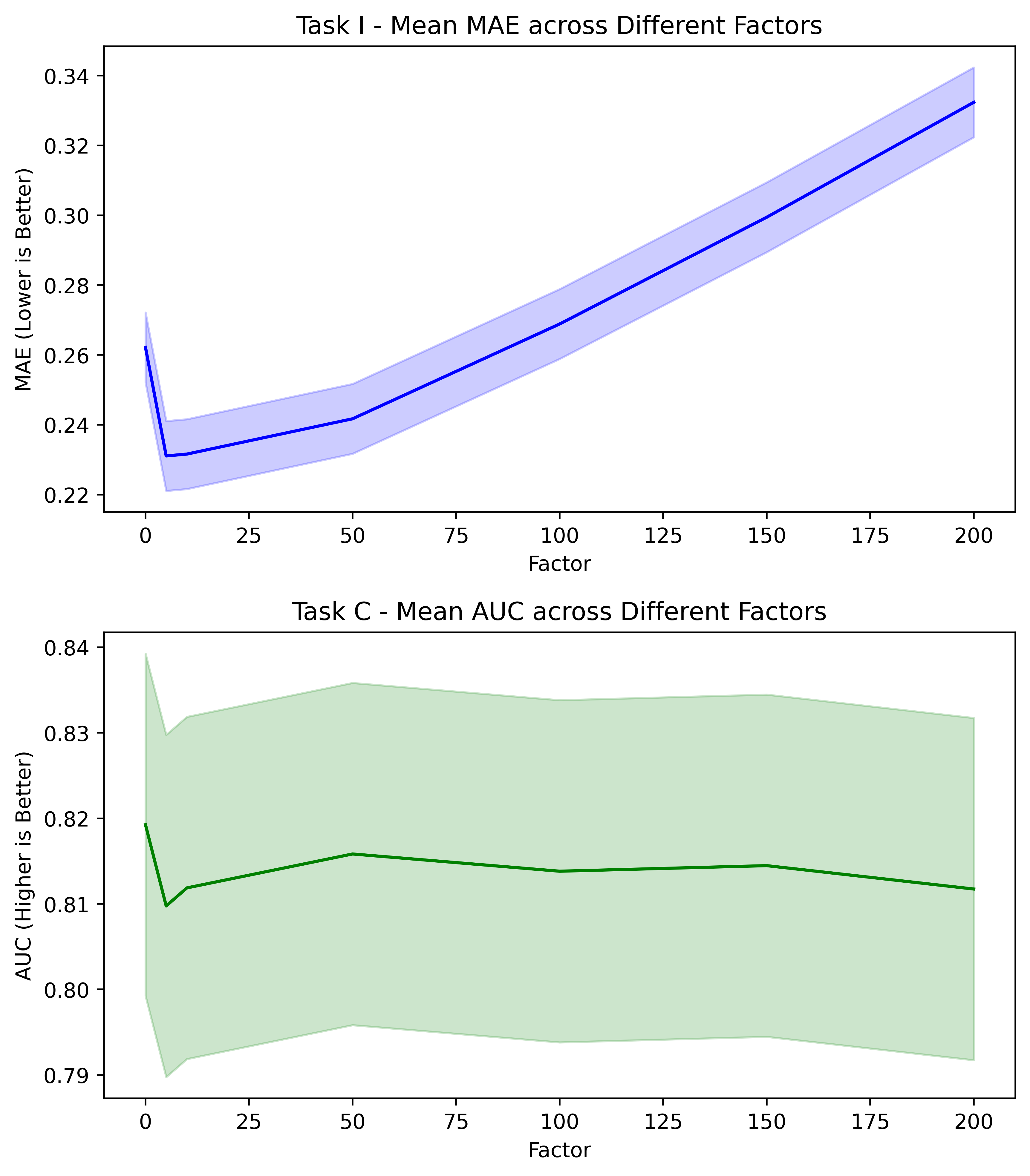}
\caption{Impact of the adjustment factor in the Physionet dataset}
\label{fig:if_exps}
\end{figure}

\section{Conclusions And Future Work}

CSAI's novelty lies in the provision of components specifically tailored to medical time-series, where the frequency and timing of data collection is highly variant, and long- and short-term correlations are pervasive. Using conditional knowledge embedding, attention mechanisms and capturing non-random missingness, CSAI outperformed established benchmarks.

CSAI's modular design enables several extensions. While CSAI currently operates within the PyPOTS ecosystem for standardised benchmarking, future clinical deployment will require privacy-preserving mechanisms. The transformer-based architecture is compatible with federated learning, where model updates are shared rather than raw data. In addition, while the current median time gap $\tau_d$ uses population-level medians, future work will explore learnable $\tau_t$ to better capture evolving clinical monitoring strategies. 

\section{Acknowledgments \& Availability}
 
All experiments were implemented on King's College London's CREATE HPC. All code and experimental details are available on the CSAI \texttt{GitHub} repository \href{https://github.com/LinglongQian/CSAI}{https://github.com/LinglongQian/CSAI}.  The open-source CSAI is immediately callable from the Python package. 

\bibliographystyle{IEEEtran}
\bibliography{sn-bibliography}

\end{document}